\documentclass[runningheads, orivec]{llncs}
\usepackage[T1]{fontenc}
\usepackage{graphicx}
\usepackage{amsmath}
\usepackage{amssymb}
\usepackage{tabularx}
\usepackage{subcaption}
\usepackage[hidelinks]{hyperref}
\usepackage{color}

\begin{document}
\title{Evolving Neural Controllers for Xpilot-AI Racing Using Neuroevolution of Augmenting Topologies}
\titlerunning{Evolving Neural Controllers for Xpilot-AI Racing Using NEAT}
%
\author{Jim O'Connor\orcidID{0009-0008-9917-5682} \and
Nicholas Lorentzen \orcidID{0009-0008-6721-726X} \and \\ 
Gary B. Parker\orcidID{0009-0001-3870-1190} \and 
Derin Gezgin\orcidID{0009-0004-0707-603X}
}
\authorrunning{O'Connor et al.}
%
\institute{Connecticut College, New London CT 06320-4125, USA \\
\email{\{joconno2,nlorentze,parker,dgezgin\}@conncoll.edu}}
\maketitle              
\begin{abstract}

This paper investigates the development of high-performance racing controllers for a newly implemented racing mode within the Xpilot-AI platform, utilizing the Neuro Evolution of Augmenting Topologies (NEAT) algorithm. By leveraging NEAT’s capability to evolve both the structure and weights of neural networks, we develop adaptive controllers that can navigate complex circuits under the challenging space simulation physics of Xpilot-AI, which includes elements such as inertia, friction, and gravity. The racing mode we introduce supports flexible circuit designs and allows for the evaluation of multiple agents in parallel, enabling efficient controller optimization across generations. Experimental results demonstrate that our evolved controllers achieve up to 32\% improvement in lap time compared to the controller's initial performance and develop effective racing strategies, such as optimal cornering and speed modulation, comparable to human-like techniques. This work illustrates NEAT's effectiveness in producing robust control strategies within demanding game environments and highlights Xpilot-AI’s potential as a rigorous testbed for competitive AI controller evolution.

\end{abstract}

\section{Introduction}

Xpilot-AI\footnote{\url{https://gitlab.com/xpilot-ai}}, developed as an extension of the classic multiplayer game Xpilot, has become a valuable tool in computational intelligence research \cite{CI-XPilot}. Originally designed as a 2D combat simulator, Xpilot integrates physics elements such as inertia, friction, and gravity, creating a simulation environment that challenges agents to manage continuous, dynamic control tasks \cite{XPilot-Extra}. In its academic adaptation, Xpilot-AI supports the design and testing of autonomous agents across various task domains, including combat and capture-the-flag modes. Its open-source, physics-driven nature allows for large-scale experimentation with computational agents, making it an accessible platform for testing a range of machine learning and neuroevolutionary methods \cite{capture-flag}. However, past applications of Xpilot-AI have primarily focused on discrete task objectives like combat and objective based path-finding, with comparatively limited exploration into continuous control tasks that emphasize speed and timing, such as racing \cite{XPILOT-PAL}.

In this study, we introduce XPRace, a racing extension to the established Xpilot-AI platform, applying the Neuro Evolution of Augmenting Topologies (NEAT) algorithm \cite{NEAT-original-paper} to evolve neural networks capable of competitive racing strategies. Competitive racing environments provide a structured platform for the development and evaluation of intelligent agents, requiring precision, adaptability, and optimized decision-making under real-time constraints \cite{racing-1}\cite{racing-2}. This work addresses the control challenges associated with high-speed, physics-driven navigation in Xpilot-AI, which adds significant complexity to tasks requiring precise agent control \cite{XPILOT-environment}. XPRace represents a novel application of NEAT in a rigorous and dynamic problem domain, leveraging its capacity to simultaneously evolve both network structure and weight parameters.

The addition of XPRace to the Xpilot-AI framework introduces a high-speed racing mode that emphasizes continuous adaptation to complex physics. Unlike traditional racing simulators, where the primary challenge is navigating a course, flying in Xpilot-AI requires agents to manage inertia and control in response to friction and gravity while including complex maps \cite{XPILOT-controller}. This presents a significant test for neuroevolutionary methods, as agents must develop not only pathfinding strategies but also efficient acceleration and braking controls to maximize speed. The complexity of Xpilot-AI’s physics model, combined with the requirements for speed optimization and handling, makes this racing task an ideal candidate for NEAT. By evolving both the topology and weights of neural networks, NEAT offers a flexible approach capable of producing high-performing agents in dynamic environments that require substantial control sophistication.

This work highlights XPRace’s potential as an experimental testbed for competitive AI, advancing the application of neuroevolutionary algorithms in domains that require continuous adaptation to challenging physics-driven tasks. By facilitating the evaluation of hundreds of agents in parallel, XPRace accelerates the optimization process, allowing for comprehensive experimentation with evolutionary approaches in competitive AI \cite{competitiveDL}. The findings demonstrate NEAT’s effectiveness in producing agents capable of real-time control optimization, supporting the broader applicability of neuroevolution in physics-intensive environments where precise control strategies are essential for success.

\section{Related Work}

Neuroevolutionary methods have been applied to various game domains, initially demonstrating success in turn-based and low-dimensional games using fixed-topology networks \cite{go-paper}. Early applications, such as Fogel’s work in evolving checkers-playing agents, laid the groundwork for evolving network weights within a fixed structure, yielding promising results in relatively simple games \cite{fogel-checkers}. However, as researchers began applying neuroevolution to continuous control domains limitations in these fixed-topology approaches became evident, particularly in environments requiring rapid adaptation and multi-faceted decision-making \cite{NE-continuousControl}. NEAT, introduced by Stanley and Miikkulainen, provided a solution to these limitations by enabling the evolution of both network weights and structure, allowing agents to develop more sophisticated architectures responsive to the complexity of the task environment \cite{NEAT-original-paper}. This innovation opened the field to applications in high-dimensional, continuous-action domains such as robotic control and racing simulations, where adaptive architectures significantly improve performance. Recent work has extended these ideas by dynamically evolving recurrent architectures using simpler evolutionary algorithms, showing strong performance on a range of control tasks with far fewer parameters \cite{extra}.

Early applications of racing AI investigated multiple controller representations in a simple racing simulator and found that neural networks trained via evolution consistently outperformed alternatives \cite{togeliusEvolve}. NEAT was applied to the Open Racing Car Simulator (TORCS), evolving controllers for aggressive (overtaking) and safe driving behaviors \cite{NEAT-racer}, which outperformed the hand-coded competition bots. Further work compared NEAT with a real-time component to the offline evolutionary strategies in TORCS \cite{NEAT-racer-2}. Another variant of NEAT, HyperNEAT was also used for evolving recurrent neural networks for autonomous driving tasks \cite{hyper}. Additionally, Covariance Matrix Adaptation Evolution Strategy is used to optimize a driving controller in TORCS, achieving competitive lap times and strategies \cite{race-es}.  

These studies show wide-range of techniques that have been used in the video game racing environments including NEAT. In contrast to prior work which focuses on traditional racing, our work applies NEAT to a space-combat simulation environment with unique control dynamics. 

\section{XPRace}

XPRace is an extension developed for Xpilot-AI platform, designed to evaluate the performance of agents in high-speed, continuous control tasks under physical constraints. Xpilot-AI introduces realistic control challenges in a 2D space combat environment. The simulation environment includes inertia, friction, and gravity, making even basic navigation a challenging task. Agents have a physical body with a mass and momentum, and must apply appropriate thrust to control its acceleration, deceleration and orientation.  

The racetrack XPRace introduces is fully enclosed by walls and the agents are not permitted to go off-track and take shortcuts. If an agent makes contact with a wall at any speed, the trial ends immediately and the agent is considered to have failed. This catastrophic nature of XPRace, pushes agents to evolve strategies that prioritize safe and efficient control policies. XPRace does not include any cumulating penalty; a single collision would result in termination. Although XPRace is capable of supporting multiple agents simultaneously, this study focuses on evolution of a single controller using NEAT. Multi-agent scenarios would include further complexity as collisions between agents would also result in termination of the race as it is fatal by the nature of Xpilot.

\begin{figure}[t]
    \centering
    \includegraphics[width=0.5\linewidth]{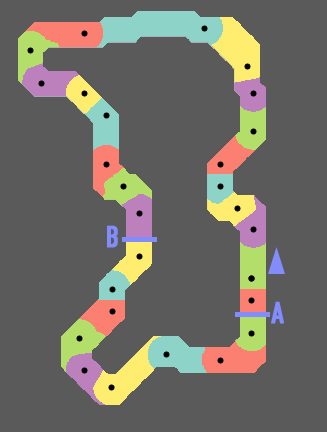}
    \caption{Map of the circuit with starting points A and B. The black dots represent the waypoints used as reference points. The colored areas indicate the vision range of the agent; the different colors are for visual distinction only and do not represent any functional difference.}
    \label{labeled-map}
\end{figure}

The racing track includes sharp corners, wide turns, and long straight-ways to expose agents to a wide range of maneuver requirements. To evaluate progress, each circuit includes a set of discrete waypoints that act as a checkpoint to track the progress of the agent. Figure~\ref{labeled-map} shows a circuit that is used in the training process. The black dots represent the waypoints, while the colored regions shows the vision range of the agent when looking ahead to find the next waypoint. It is important to note that the agent is not enforced to exactly hit these waypoints to complete the navigation. Agents are free to develop their own optimal racing strategy and rewarded for the completion of the track and the completion time. In fact, the default path produced by the waypoints is suboptimal in terms of speed and efficiency. Agents must learn to exploit the physics of the environment and produce more effective strategies. 

\subsection{Input/Output Parameters of the Racing Agents}

The successful evolution of effective controllers in this environment depends on providing each agent with sufficient input data to make real-time navigational decisions and plan future actions. The agent has two output variables: a scalar representing the desired steering angle and a second scalar representing the thrust component. Each agent receives 23 input parameters as follows: \textit{Ship Speed}, speed of the agent divided by 20.0 to normalize within the 0.0 to 1.0 range. \textit{Wall Track}, indicating the distance to the wall along the current track direction. \textit{Angle Diff Tracking} representing the angular difference between the ship’s heading and the track direction. \textit{Closest Wall} provides the distance to the nearest wall, regardless of direction. \textit{Angle Diff Closest} measures the angular difference between the ship’s heading and the closest wall. Additionally, eight parameters represent distances to the walls at 15 and 30 degrees to the left and right of the current heading, as well as the cardinal directions. Further input parameters include \textit{TT Tracking} which provides time-to-collision with the wall along the track direction, and \textit{TT Retro Point} which provides time in steps to the last point where retrograde thrust could prevent a collision. Additionally \textit{Last Thrust} records the agent’s last thrust output and \textit{Last Turn} records the previous steering angle. The final six parameters describe the angles and distances to the next three waypoints relative to the current ship heading, enabling the agent to anticipate upcoming track features. Table~\ref{tab:params} summarizes the input parameters of the racing agents.

\begin{table}[!t]
\centering
\caption{The input parameters for the evolved controller.}
\begin{tabularx}{\textwidth}{|p{4cm}|X|}
  \hline
  \textbf{Parameter} & \textbf{Description} \\
  \hline
  Ship Speed (scaled) & Normalized ship speed value (0.0 to 1.0). \\
  \hline
  Wall Track & Distance to wall in the track direction. \\
  \hline
  Angle Diff \& Tracking & Angular difference between the current heading and the track direction. \\
  \hline
  Closest Wall \& Angle Diff & Distance to the nearest wall in any direction and the angular difference to that wall. \\
  \hline
  Wall Distances (0°, 180°, ±15°, ±30°, ±90°) & Distances to walls in cardinal and angled directions around the ship. \\
  \hline
  TT Tracking \& TT Retro Point & Time to collision with the wall in the track direction and time to last retrograde thrust point. \\
  \hline
  Last Thrust \& Last Turn & Previous thrust and steering angle output values. \\
  \hline
  Waypoint Angles \& Distances & Angles and distances to the next three waypoints relative to the current ship heading. \\
  \hline
\end{tabularx}
\label{tab:params}
\end{table}

\section{NEAT Applied to XPRace}

\subsection{Defining A Fitness Function}
\begin{figure}[!t]
    \centering
    \includegraphics[width=0.5\linewidth]{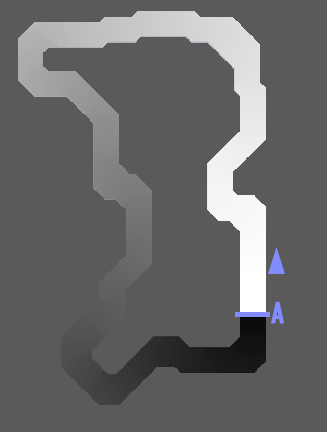}
    \caption{Example visualization of the completion function for starting point A. The completion scales from 0-100\% from light to dark. This is used in Equation \eqref{fitness} in order to reward the agent for a higher completion.}
    \label{distinct-start}
\end{figure}

The evolution of agents capable of completing a racing circuit and optimizing their lap times requires an effective fitness function. The fitness function is structured to prioritize course completion before optimizing speed, with agents receiving fitness based on both factors sequentially. Agents earn an initial fitness score by achieving incremental completion levels (0-100\%), computed using waypoint data for each circuit which can be seen in Figure \ref{distinct-start}. The waypoints are only for calculation purposes while the agent does not have to exactly hit them. This completion metric is raised to the power of 1.1 to emphasize early improvements in progress and encourage agents to explore efficient navigation strategies before focusing on speed optimization.

To encourage speed optimization, a frame-level bonus is accumulated during episodes when the agent maintains a speed above 1.0, using a scaled power function 
$(\text{speed}^{1.1}/250)$ that is normalized between 0 and 1. This bonus is awarded to the agent when it is alive and making progress, and it is capped at a maximum of 50.00 in the final fitness calculation to prevent over-emphasis on raw speed. Additionally, a time metric is calculated based on a baseline lap time from pre-trial runs. The time-based fitness is computed by subtracting the agent’s time from this baseline and adding 75, then raising the sum to the power of 1.2 to increase the reward for further time reductions.

When agents are evaluated across multiple starting configurations, fitness scores for each configuration are summed to provide the final fitness score. Each evaluation period includes a single lap attempt per starting point, ending upon course completion, collision, halted progress (3 seconds), or exceeding the time limit. The final fitness equation is defined as:

\begin{equation}
    Fit_i=C_{i}^{1.1}+Min(B_i, 50) + \begin{cases}
    (T_i + 75)^{1.2}, & \text{if $T_i$ exists}.\\
    0, & \text{otherwise}.
  \end{cases}
  \label{fitness}
\end{equation}
where $i$ represents the starting point the agent is evaluated on (A or B), $C_i$ denotes completion percentage on starting point $i$, $B_i$ denotes the speed bonus on starting point $i$, and $T_i$ represents the difference between the agent's lap time and the target time. Adding 75 to $T_i$ ensures a fitness advantage for agents completing at least one circuit configuration. 

\subsection{Mitigating First-Turn Bias}
A challenge in evolving navigational agents is the potential bias toward the initial segment of the course. This can result in overfitting to the starting area, delaying the first complete lap. To mitigate this, agents are evaluated from two distinct starting points, denoted A and B in Figure \ref{labeled-map}. Notably, both of these starting points are positioned on straight segments of the course but differ in initial turn direction. This configuration balances early-stage evolution, preventing the dominance of one-sided navigation strategies. Fitness values from both starting points are combined to form the overall fitness, reducing potential bias in evolving the population.

\begin{table}[!b]
\centering
\caption{Summary of the First and Final Trial Durations and Generation Number for Each Starting Point A and B}
\begin{tabular}{|c|c|c|c|c|c|c|}
\hline
\textbf{Trial} &
\multicolumn{3}{c|}{\textbf{Point A}} &
\multicolumn{3}{c|}{\textbf{Point B}} \\
\cline{2-7}
& \textbf{Start Time} & \textbf{End Time} & \textbf{Improvement} 
& \textbf{Start Time} & \textbf{End Time} & \textbf{Improvement} \\
\hline
1 & \begin{tabular}[c]{@{}c@{}}55.766\\(Gen. 389)\end{tabular} 
  & \begin{tabular}[c]{@{}c@{}}39.370\\(Gen. 1000)\end{tabular} 
  & 29.4\% 
  & \begin{tabular}[c]{@{}c@{}}56.108\\(Gen. 371)\end{tabular} 
  & \begin{tabular}[c]{@{}c@{}}41.675\\(Gen. 1000)\end{tabular} 
  & 25.7\% \\
\hline
2 & \begin{tabular}[c]{@{}c@{}}63.352\\(Gen. 226)\end{tabular} 
  & \begin{tabular}[c]{@{}c@{}}44.216\\(Gen. 1000)\end{tabular} 
  & 30.2\% 
  & \begin{tabular}[c]{@{}c@{}}64.528\\(Gen. 230)\end{tabular} 
  & \begin{tabular}[c]{@{}c@{}}47.972\\(Gen. 1000)\end{tabular} 
  & 25.7\% \\
\hline
3 & \begin{tabular}[c]{@{}c@{}}56.940\\(Gen. 166)\end{tabular} 
  & \begin{tabular}[c]{@{}c@{}}42.552\\(Gen. 1000)\end{tabular} 
  & 25.3\% 
  & \begin{tabular}[c]{@{}c@{}}64.363\\(Gen. 132)\end{tabular} 
  & \begin{tabular}[c]{@{}c@{}}47.500\\(Gen. 1000)\end{tabular} 
  & 26.2\% \\
\hline
4 & \begin{tabular}[c]{@{}c@{}}61.835\\(Gen. 307)\end{tabular} 
  & \begin{tabular}[c]{@{}c@{}}41.524\\(Gen. 998)\end{tabular} 
  & 32.8\% 
  & \begin{tabular}[c]{@{}c@{}}53.302\\(Gen. 224)\end{tabular} 
  & \begin{tabular}[c]{@{}c@{}}45.778\\(Gen. 1000)\end{tabular} 
  & 14.1\% \\
\hline
5 & \begin{tabular}[c]{@{}c@{}}53.368\\(Gen. 426)\end{tabular} 
  & \begin{tabular}[c]{@{}c@{}}44.754\\(Gen. 1000)\end{tabular} 
  & 16.1\% 
  & \begin{tabular}[c]{@{}c@{}}57.288\\(Gen. 422)\end{tabular} 
  & \begin{tabular}[c]{@{}c@{}}46.651\\(Gen. 1000)\end{tabular} 
  & 18.6\% \\
\hline
\textbf{Mean} & 58.252 & 42.483 & 27.1\% & 59.118 & 45.915 & 22.3\% \\
\hline
\end{tabular}
\label{tab:merged-comparison}
\end{table}

\subsection{Trial Configuration}
A population size of 200 was selected to optimize training efficiency and performance. Initially, agents are structured with two hidden layers and random partial connections. A stagnation limit of 50 generations is applied to allow new species sufficient time to improve fitness but to also permit extinction when stagnation occurs. The stagnation limit prevents species from persisting indefinitely if they fail to show improvement, get stuck in a local-optimum. If a species does not show any progress for 50 generations, it is removed to ensure computational resources are used for evolving more adaptive controllers. Elitism was set at 4 to prevent over-concentration of any single strategy, ensuring that a minimum of three additional species are retained in the population to foster diversity.

\begin{figure*}[!ht]
    \centerline{
        \includegraphics[width=1\textwidth]{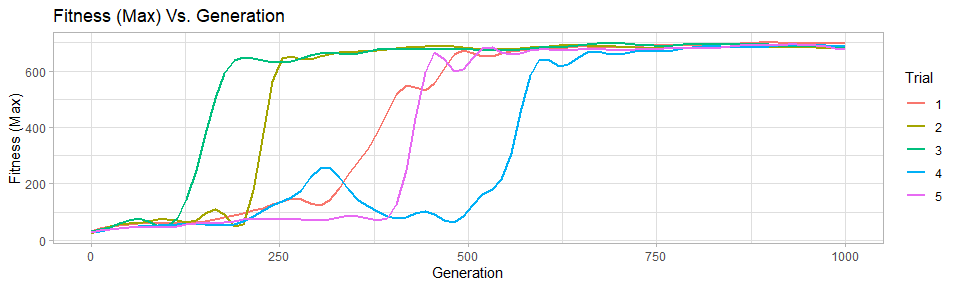}}
        \caption{Fitness graph across 1,000 generations. Each line represents the highest fitness score in a given trial.}
        \label{fitneessfig}
\vspace{1em}
    \begin{minipage}[t]{0.5\textwidth}
        \centering
        \includegraphics[width=\textwidth]{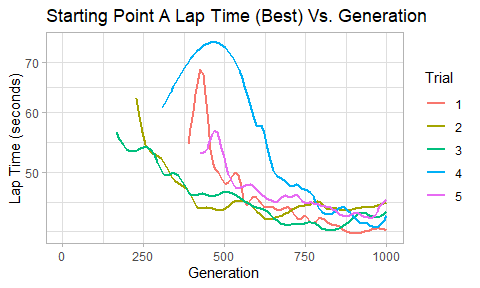}
        \caption{Best lap times for starting point A.}
        \label{pointAfig}
    \end{minipage}
    \begin{minipage}[t]{0.5\textwidth}
        \centering
        \includegraphics[width=\textwidth]{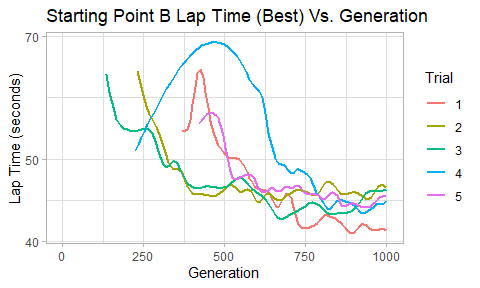}
        \caption{Best lap times for starting point B.}
        \label{pointBfig}
    \end{minipage}
\end{figure*}

\section{Results}

This training procedure was repeated for 5 individual trials for starting points A and B in Figure~\ref{distinct-start}. In all five trials, agents successfully completed the course within the 1,000 generation limit imposed on both starting points.  The first instances of agents completing both starting configurations appeared between generation 166 (Trial 3 in starting point A) and generation 426 (Trial 5 in starting point A). The results in Figure~\ref{fitneessfig} show a rapid increase in fitness at different positions in the training, with all trials eventually converging to a high-performing solution. The variation in the timing of the fitness increase shows that different evolutionary paths lead to successful racing strategies. The summary of the first and final completion times for each starting point is presented in Table~\ref{tab:merged-comparison}

\subsection{Lap Time Optimization}

Agents showed steady improvements across all trials. For starting point A in Figure~\ref{pointAfig}, initial times averaged 58.252 seconds across trials, with the final best times ranging from 39.370 to 44.754 seconds, averaging 42.483 seconds, a 27.1\% improvement compared to the initial average. For starting point B in Figure~\ref{pointBfig}, initial times averaged 59.118 seconds, with best times ranging from 41.675 to 47.972 seconds, averaging 45.915 seconds— a 22.3\% improvement compared to the initial average. Combined times for trials (sum of starting points A and B for each trial) initially averaged at 117.370 seconds, improving to an average best time of 88.398 seconds across trials. The largest improvement was 32.8\% (Trial 4 in starting point A), while the smallest was 14.1\% (Trial 4 in starting point B).

\begin{figure}[!t]
    \centering
    \includegraphics[width=0.5\linewidth]{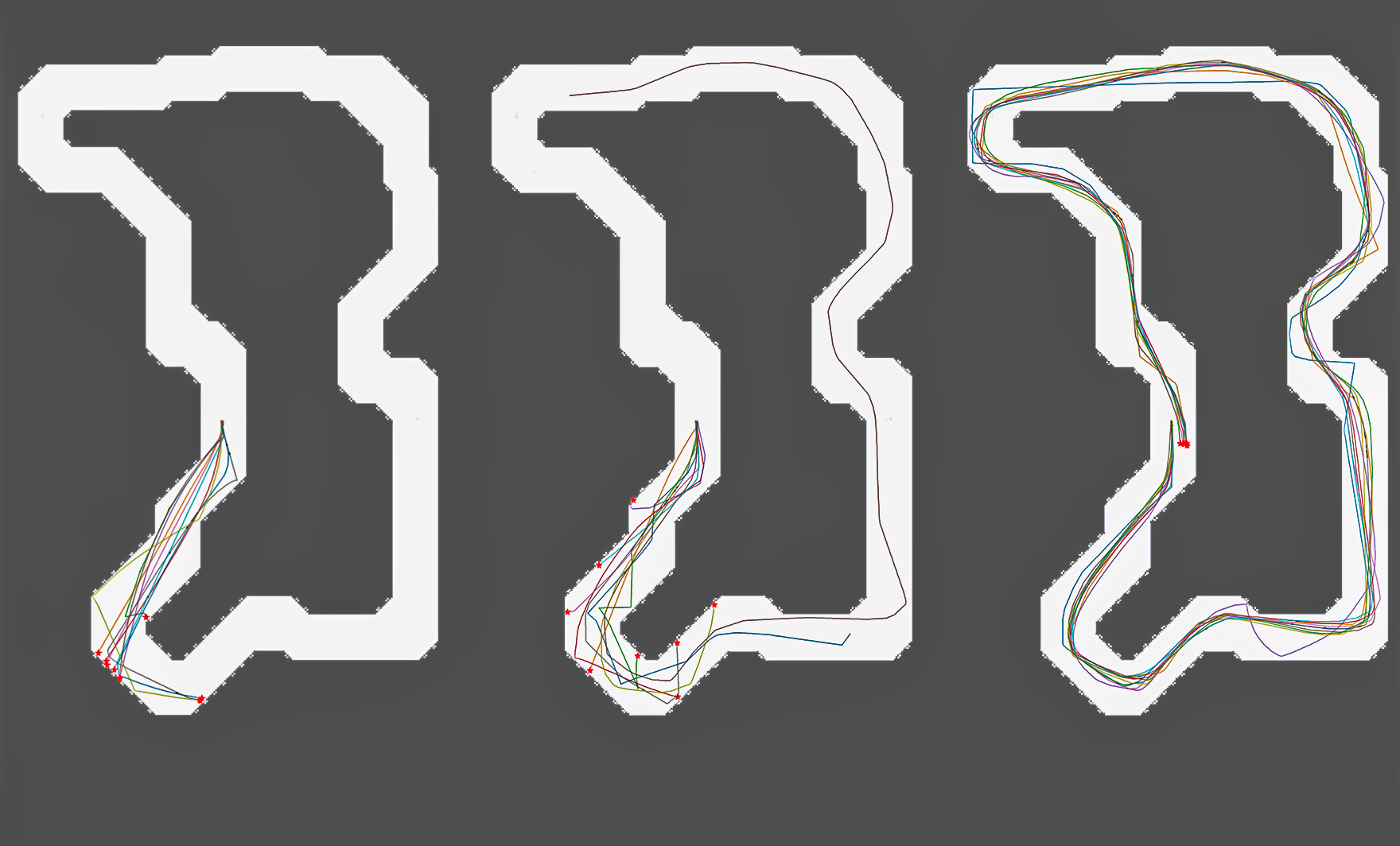}
    \caption{The racing lines from starting point B of the top 10 agents from different species in 3 sample generations. Red dots indicate the location where an agent crashed or finished the circuit.}
    \label{racinglinesfig}
\end{figure}

We observed that, agents evolved strategies resembling those observed in real-world racing. Analysis of top-performing agents across trials in Figure \ref{racinglinesfig} reveals tendencies to optimize paths. The racing paths of the most-evolved, right-most, controllers clearly shows the “Outside-Inside-Outside” cornering techniques while dynamically adjusting the speed to maintain stability and minimize time loss in turns. NEAT's speciation facilitated slight variations in line-taking strategies among the top agents.

\section{Conclusion}

This study has demonstrated the viability and effectiveness of applying the Neuro Evolution of Augmenting Topologies (NEAT) algorithm to evolve racing controllers within a challenging 2D, space physics-based environment. By leveraging the complex dynamics of the Xpilot-AI framework, including inertia, friction, and variable gravity, we successfully evolved agents capable of competitive racing performance in a newly introduced racing mode. In all 5 trials, the agents successfully learned to complete a circuit within the first 430 generations, completing the circuit as early as generation 132. The evolved controllers consistently improved lap times by up to 32\% across generations, with some agents displaying racing strategies that emulate human approaches, such as optimal cornering and adaptive throttle modulation. These findings highlight NEAT’s capacity to optimize neural architectures in dynamic control tasks, especially in scenarios that demand rapid adaptation to real-time environmental feedback.

\subsection{Future Work}

Moving forward, we aim to expand upon the insights gained in this study by exploring more advanced neuroevolutionary techniques, such as ES-HyperNEAT and other NEAT derivatives, to evolve controllers capable of managing the unique challenges of the Xpilot-AI environment. These methods offer the potential to handle higher-dimensional input spaces and more complex control strategies, thereby improving the generality and robustness of the evolved agents. Additionally, while our study focuses on training a single agent on a single track with two distinct starting points, the evolved agents can be trained and tested on a variety of tracks to increase the adaptability of the agent. It is also possible to extend our single-agent setting to multi-agent racing scenarios.

\bibliographystyle{splncs04}
\bibliography{references}

\end{document}